\documentclass[report]{tufa}

\usepackage{float}
\AddToHook{cmd/appendix/after}{\floatplacement{figure}{H}\floatplacement{table}{H}}
\usetikzlibrary{arrows.meta,backgrounds,fit,positioning}

\setcitestyle{authoryear,square,comma}
\newcommand{\inlinebreak}{\allowbreak}
\newcommand{\layoutbreak}{}

\makeatletter
\def\NAT@sort{0}
\def\NAT@cmprs{0}
\makeatother

\newcommand{\papertitle}{Context--Answer Metamodels: Forecasting LLM Activations Before the First Answer Token}
\newcommand{\firstauthorname}{Tommy He}
\newcommand{\firstauthoraffiliation}{Tufa Labs}
\newcommand{\secondauthorname}{Thomas Jiralerspong}
\newcommand{\secondauthoraffiliation}{Mila, Université de Montréal}

\tufasetup{
  title={\papertitle},
  short-title={Context–Answer Metamodels},
  authors={\firstauthorname\textsuperscript{1}\quad\secondauthorname\textsuperscript{2}},
  pdf-authors={\firstauthorname\ and \secondauthorname},
  affiliation={\textsuperscript{1}\firstauthoraffiliation\quad\textsuperscript{2}\secondauthoraffiliation},
  date={4 November 2026},
  contact={tommy@tufalabs.ai},
  code-url={https://github.com/Tufalabs/context-answer-metamodels},
}

\begin{document}

\begin{abstract}
Under stochastic decoding, a single context can elicit many different answers from a language model. Characterizing this variation typically requires many expensive autoregressive rollouts, each revealing only one realization. We ask whether the activations produced while reading the context can directly forecast the model's rollout activations.  We train \emph{context--answer metamodels} that map a context activation sequence to point or distribution forecasts of mean-pooled answer activations. We find that (i) the four metamodel families we study forecast mean-pooled answer activations with high fidelity, and the distribution metamodels also capture rollout variability; (ii) performance improves as the training set grows to 500,000 prompts, and the metamodels generalize zero-shot to out-of-domain datasets; (iii) their forecasts retain behavior-predictive signal; and (iv) context activations from Qwen2.5 predict mean-pooled answer activations from Qwen3.5 and Gemma 4, suggesting that activation forecasts can transfer across models. Together, these results suggest that context activation traces can support scalable forecasts of behaviors in LLM answers before the first answer token is generated.
\end{abstract}

\maketitle
\section{Introduction}
How much of a language model's future output is predictable before it
emits its first answer token?  Under stochastic decoding, a single context can produce different answer trajectories even when the model and decoding configuration are held fixed.
Estimating this variability normally requires sampling multiple expensive
rollouts.  We ask whether we can forecast it directly from context activations.
Such a forecast is useful in applications where a single completion is an unreliable indicator of a model's possible behavior, such as pre-generation rare-behavior screening and pass@$k$ estimation.
Behavior screening often samples many answers and scores them with a probe
or another language model.  This is costly for rare events, and it provides no
warning before the sampled behavior occurs.
Rather than modeling combinatorially many complete token sequences directly, we forecast their answer representations.

Prior work predicts later tokens and hidden states from earlier activations \citep{pal2023future}, while other work uses context activations to predict answer attributes, hallucination risk, and answer correctness \citep{dong2025planning,ji2024hallucination,moreno2025noanswer}.
Most closely related, \citet{jiralerspong2026affine} fit linear and nonlinear
context--answer maps on 500,000 training prompts, with one sampled answer per
prompt. We study point and distribution
forecasts under repeated-rollout supervision and evaluate both.

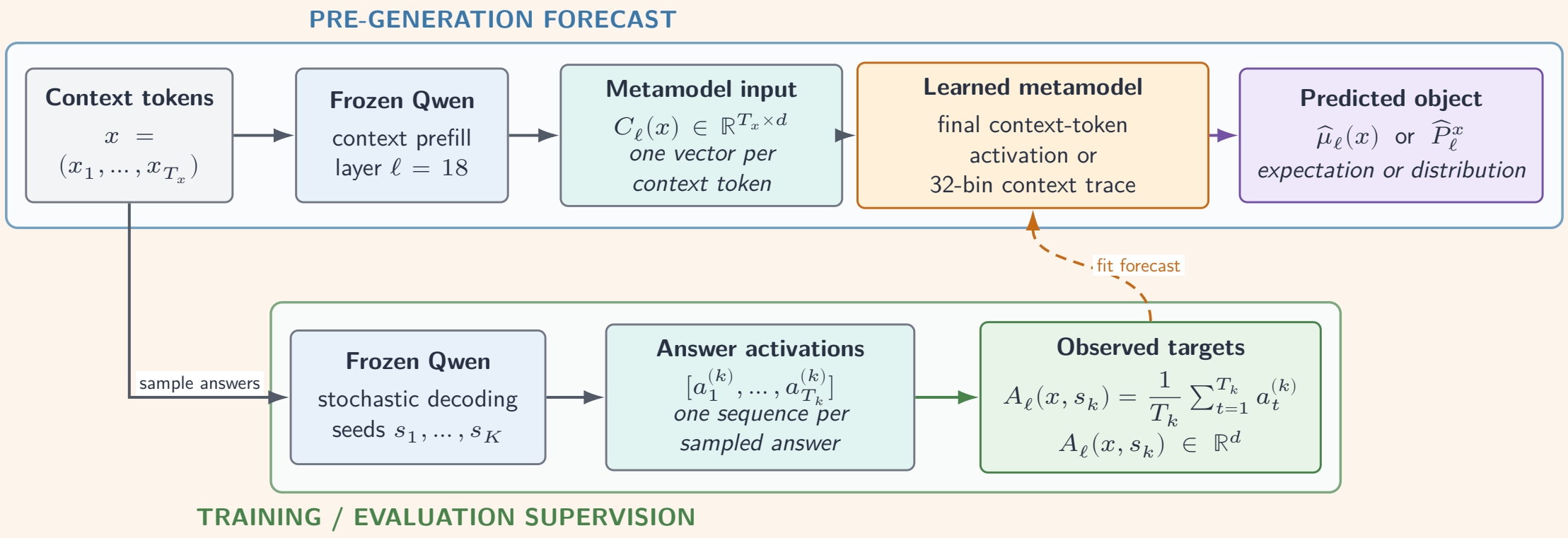
\begin{figure}[t]
    \centering
    \resizebox{0.94\linewidth}{!}{\definecolor{schematicPrompt}{HTML}{F3F4F6}
\definecolor{schematicFrozen}{HTML}{E8F1FA}
\definecolor{schematicTrace}{HTML}{E2F4F1}
\definecolor{schematicLearned}{HTML}{FFF0D9}
\definecolor{schematicForecast}{HTML}{EEE8FA}
\definecolor{schematicTarget}{HTML}{E8F4E7}
\definecolor{schematicInk}{HTML}{273142}
\definecolor{schematicBlue}{HTML}{3978A8}
\definecolor{schematicOrange}{HTML}{C66A16}
\definecolor{schematicGreen}{HTML}{47834A}
\definecolor{schematicPurple}{HTML}{7453A6}

\begin{tikzpicture}[
    x=1cm,
    y=1cm,
    font=\sffamily\scriptsize,
    text=schematicInk,
    box/.style={
        draw=schematicInk!65,
        line width=0.55pt,
        rounded corners=2pt,
        align=center,
        inner sep=4pt,
        minimum height=1.25cm
    },
    arrow/.style={-{Latex[length=2mm,width=1.35mm]}, line width=0.75pt,
                  draw=schematicInk!80},
    supervision/.style={-{Latex[length=2mm,width=1.35mm]}, line width=0.7pt,
                        densely dashed, draw=schematicOrange},
    region/.style={rounded corners=3pt, line width=0.65pt, inner sep=5pt}
]

\node[box, fill=schematicPrompt, text width=1.65cm] (prompt) at (0,0) {
    \textbf{Context tokens}\\[2pt]
    $x=(x_1,\ldots,x_{T_x})$
};

\node[box, fill=schematicFrozen, text width=1.70cm] (prefill) at (2.55,0) {
    \textbf{Frozen Qwen}\\[2pt]
    context prefill\\
    layer $\ell=18$
};

\node[box, fill=schematicTrace, text width=2.35cm] (trace) at (5.35,0) {
    \textbf{Metamodel input}\\[2pt]
    $C_\ell(x)\in\mathbb{R}^{T_x\times d}$\\[-1pt]
    \emph{one vector per context token}
};

\node[box, fill=schematicLearned, draw=schematicOrange,
      text width=3.00cm] (metamodel) at (8.45,0) {
    \textbf{Learned metamodel}\\[2pt]
    final context-token\\
    activation or\\
    32-bin context trace
};

\node[box, fill=schematicForecast, draw=schematicPurple,
      text width=2.55cm] (forecast) at (11.80,0) {
    \textbf{Predicted object}\\[2pt]
    $\widehat\mu_\ell(x)\;\text{or}\;\widehat P_\ell^x$\\[1pt]
    \emph{expectation or distribution}
};

\draw[arrow] (prompt) -- (prefill);
\draw[arrow] (prefill) -- (trace);
\draw[arrow] (trace) -- (metamodel);
\draw[arrow, draw=schematicPurple] (metamodel) -- (forecast);

\node[box, fill=schematicFrozen, text width=2.10cm] (decode) at (2.70,-2.45) {
    \textbf{Frozen Qwen}\\[2pt]
    stochastic decoding\\
    seeds $s_1,\ldots,s_K$
};

\node[box, fill=schematicTrace, text width=2.60cm] (answertrace) at (5.90,-2.45) {
    \textbf{Answer activations}\\[2pt]
    $[a_1^{(k)},\ldots,a_{T_k}^{(k)}]$\\[-1pt]
    \emph{one sequence per sampled answer}
};

\node[box, fill=schematicTarget, draw=schematicGreen,
      text width=2.90cm] (targets) at (9.55,-2.45) {
    \textbf{Observed targets}\\[2pt]
    $A_\ell(x,s_k)=\dfrac{1}{T_k}\sum_{t=1}^{T_k}a_t^{(k)}$\\
    $A_\ell(x,s_k)\in\mathbb{R}^{d}$
};

\draw[arrow] (prompt.south) |- node[pos=0.72, above, fill=white,
    inner sep=1pt, font=\sffamily\tiny]
    {sample answers} (decode.west);
\draw[arrow] (decode) -- (answertrace);
\draw[arrow, draw=schematicGreen] (answertrace) -- (targets);
\draw[supervision] (targets.north) to[out=90,in=-90]
    node[right, fill=white, inner sep=1pt, font=\sffamily\tiny,
         text=schematicOrange]
    {fit forecast} (metamodel.south);

\begin{scope}[on background layer]
    \node[region, draw=schematicBlue!70, fill=schematicBlue!3,
          fit=(prompt)(prefill)(trace)(metamodel)(forecast),
          label={[font=\sffamily\scriptsize\bfseries,text=schematicBlue]
          above left:PRE-GENERATION FORECAST}] {};
    \node[region, draw=schematicGreen!70, fill=schematicGreen!3,
          fit=(decode)(answertrace)(targets),
          label={[font=\sffamily\scriptsize\bfseries,text=schematicGreen]
          below left:TRAINING / EVALUATION SUPERVISION}] {};
\end{scope}

\end{tikzpicture}}
    \caption{\textbf{Forecasting a future activation distribution.}
    The metamodel receives context activations from a language
    model and produces either a point forecast of the expected mean-pooled
    answer activation (linear ridge or MLP) or a distribution forecast over
    mean-pooled answer activations (Gaussian or flow). Multiple decoding seeds provide training
    and evaluation targets.}
    \label{fig:task}
\end{figure}

We train \emph{context--answer metamodels}\footnote{Code repository: \url{https://github.com/Tufalabs/context-answer-metamodels}.} on context activation traces and repeated answer activation traces from rollouts of a language model.
\emph{Point} metamodels forecast the expectation of the mean-pooled answer activation; \emph{distribution}
metamodels forecast its rollout-induced distribution.
Figure~\ref{fig:task} summarizes the training and evaluation procedure.  We additionally test
whether the forecasts preserve behavior-predictive directions present in real
answer activations.

Our contributions are:
\begin{itemize}
    \item We compare linear ridge and MLP predictors of the conditional mean with
    Gaussian and flow models of rollout variability.
    \item We show continued scaling through 500,000 training prompts and zero-shot out-of-domain transfer from LMSYS to WeirdChat and IFEval.
    \item We show that forecasted answer activations retain behavior-predictive signal.
    \item We show that context activations from Qwen2.5 can forecast Qwen3.5 and Gemma 4 answer activations, retaining most of the accuracy of matched target-model conditioners at scale.
\end{itemize}

\section{Methodology}
\label{sec:method}

\subsection{Problem formulation}

Let $M$ be a frozen autoregressive language model with hidden dimension $d$,
$D$ a fixed stochastic decoding configuration, and $x$ an input context (prompt) with token
length $T_x$.  Let $s\sim\mathbb P_D$ denote a pseudorandom seed.  Prefill at the output of decoder block
$\ell$ yields a context activation trace
$C_\ell(x) \in \mathbb{R}^{T_x \times d}$.  A rollout under $D$ yields a
variable-length answer activation trace that we mean-pool
into $A_\ell(x,s) \in \mathbb{R}^d$
\citep{wang2026truth,der2026turnaveraged}.  This stochastic decoding process
defines both a conditional expectation and a conditional law:
\begin{equation}
\begin{aligned}
    \mu_\ell(x)
      &:= \mathbb E_{s\sim\mathbb P_D}\!\left[A_\ell(x,s)\mid x\right]\\
    P_\ell^x
      &:= \operatorname{Law}_{M,D}\!\left(A_\ell(x,s)\mid x\right).
\end{aligned}
\end{equation}
We accordingly study two prediction targets: a \emph{point metamodel}
models the expectation $\mu_\ell$, and a \emph{distribution metamodel}
models the whole rollout-induced law, $P_\ell^x$.  We write
$\widehat\mu_\ell(x)$ and $\widehat P_\ell^x$ for the corresponding learned
forecasts, appending a model-family subscript only when needed.

\subsection{Four metamodel families}

We train two point metamodels (linear ridge and MLP) and two distribution
metamodels (Gaussian and flow). We select the former by
validation $R^2$ and the latter by validation energy score.
For inputs, the linear ridge model takes the final context-token
activation, whereas the MLP and flow take a 32-bin context trace (the
Gaussian extends a frozen MLP). Adaptive average pooling forms $B=32$
contiguous, approximately equal-sized bins that may overlap, reusing tokens
when $T_x<32$.

We pool the bin averages $r_1,\ldots,r_B\in\mathbb{R}^d$
with a learned query $q$:
\begin{equation}
    \alpha_j=\operatorname{softmax}_j
      \left(\frac{q^\top\operatorname{LN}(r_j)}{\sqrt d}\right),
    \qquad z=\sum_{j=1}^{B}\alpha_j r_j .
\end{equation}
We use single-query, query-only attention pooling: $q$ is one learned global query, the normalized bins directly compute its attention scores, and the unprojected bins are the
values. This pooling is the first learned stage: $z$ is passed directly to the
downstream MLP or flow, and $q$ is trained jointly with it.

\paragraph{Linear ridge point metamodel.}
The linear ridge forecast is $\widehat\mu_{\ell,\mathrm{lin}}(x)=W\widetilde h_{\mathrm{last}}+b$,
where $\widetilde h_{\mathrm{last}}$ is the final context-token activation,
coordinatewise standardized using training statistics. We fit this
full-dimensional map to empirical rollout means in closed form, selecting
the ridge penalty $\lambda$ by validation $R^2$.
Appendix Figure~\ref{fig:linear-point-representations} ablates the choice of input representation.

\paragraph{MLP point metamodel.}
The MLP forecast $\widehat\mu_{\ell,\mathrm{MLP}}(x)$ maps $z$ through a three-layer
$4{,}096{\rightarrow}\inlinebreak8{,}192{\rightarrow}\inlinebreak8{,}192{\rightarrow}\inlinebreak4{,}096$ GELU MLP
(134.2M parameters) and minimizes MSE against each prompt's empirical mean over
training rollouts.  We select the final model with the highest validation $R^2$.  Appendix
Figure~\ref{fig:point-mlp-conditioners} ablates the choice of input representation.

\paragraph{Gaussian distribution metamodel.}
This model extends a frozen MLP point map as
$\widehat P_{\ell,\mathrm G}^x=\mathcal N(\widehat\mu_{\ell,\mathrm{MLP}}(x),\inlinebreak
\kappa^2\gamma(z)^2\Sigma)$.  We estimate $\Sigma$ from within-prompt residuals
$A_\ell(x,s)-\overline A_\ell(x)$, where $\overline A_\ell(x)$ is the empirical
rollout mean. Candidate covariances are isotropic, diagonal, or
low-rank-plus-diagonal.  The scale is global ($\gamma(z)=1$) or prompt-dependent.
For the latter, $\gamma(z)$ is a normalized scale derived from a two-layer
$4{,}096{\rightarrow}256{\rightarrow}1$ SiLU head's log-variance prediction
(Appendix~\ref{app:reproducibility}). Covariance family, rank, scale mode, and
the scalar multiplier $\kappa$ are selected by validation energy score.

\paragraph{Flow distribution metamodel.}
We use conditional flow matching with a linear path
\citep{lipman2023flow}. Integrating $v_\theta(y_t,t,z)$ from a Gaussian base
draw $\varepsilon$ at $t=0$ to $t=1$ produces a sample
$\widetilde A_\ell\sim\widehat P_{\ell,\mathrm F}^x$.
For each rollout target $y$, base draw $\varepsilon$, and $t\sim\mathcal U[0,1]$,
we set $y_t=(1-t)\varepsilon+ty$ and minimize
\begin{equation}
    \mathcal L_{\mathrm{FM}}(\theta)
      =\mathbb E\left[
        \left\|v_\theta(y_t,t,z)-(y-\varepsilon)\right\|_2^2
      \right].
\end{equation}
The vector field $v_\theta$ has width 4,096, sinusoidal time embeddings, and eight
conditioned residual blocks (352.8M parameters). At inference, 64 Gaussian
base draws are integrated for 32 Euler steps. We select the
final model with the lowest validation energy score. Appendix
Figure~\ref{fig:flow-conditioners} ablates the choice of input representation.

\subsection{Experimental setup}
\label{sec:setup}

Our primary study uses decoder block 18 of Qwen3.5-9B
\citep{qwen2026qwen35}, selected in Appendix Figure~\ref{fig:layer-selection}.
All layer indices are zero-indexed and refer to raw decoder-block outputs.
LMSYS-Chat-1M \citep{zheng2024lmsyschat1m} provides 500,000/1,000/3,000 train/validation/test prompts, each with four independently seeded rollouts. The rollout-count ablation uses 100,000 training prompts, the same validation/test sizes, and a common 16-rollout evaluation target.
Out-of-domain (OOD) evaluation uses all 2,660 WeirdChat prompts
\citep{chowdhury2026weirdchat} and 541 IFEval prompts \citep{zhou2023ifeval},
with frozen LMSYS metamodels and four observed rollouts per prompt.
We evaluate the linear ridge and MLP forecasts and the flow mean using pooled
multivariate $R^2$ against the empirical rollout mean (``mean $R^2$'').
The Gaussian's analytic mean equals the MLP forecast. For the Gaussian and flow,
we add the energy score \citep{gneiting2007strictly} divided by $\sqrt d$, Spearman
correlation between predicted and observed prompt-level variance, and the ratio
of generated to observed variance (definitions and secondary metrics in
Appendix~\ref{app:metrics}).
For cross-model forecasting, Qwen2.5-7B-Instruct layer~19 context activations
\citep{qwen2024qwen25} predict Qwen3.5-9B layer~18 and Gemma-4-12B-it layer~47
answer activations \citep{gemma2026gemma4}, with matched target-model conditioners
as controls. Layers 19 and 47 are fixed from earlier validation-selected
final context-token activation sweeps (Appendix~\ref{app:cross-model}).
Appendix~\ref{app:reproducibility} gives the full experimental settings.

\section{Results}
\label{sec:results}

\subsection{Context activation traces forecast answer activations at scale and generalize OOD}

\begin{figure}
    \centering
    \includegraphics[width=0.94\linewidth]{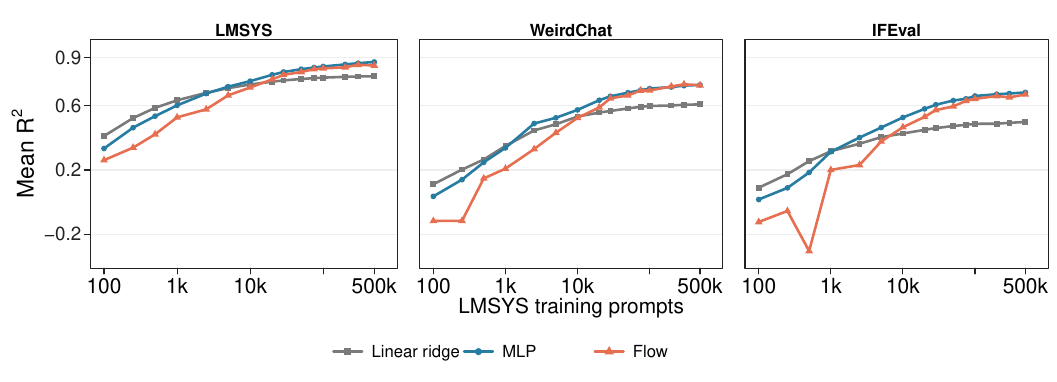}
    \caption{%
        \textbf{Prompt coverage drives forecast quality through the largest
        tested scale, in and out of domain.}
        Columns correspond to evaluation sets and show mean $R^2$ for the linear ridge, MLP,
        and flow metamodels. All metamodels are trained only on
        LMSYS; WeirdChat (2,660 prompts) and IFEval (541 prompts) are zero-shot, with four
        rollouts per prompt.  The linear ridge metamodel uses the final context-token activation and
        the neural models use the 32-bin context trace. Other endpoint
        metrics are in
        Table~\ref{tab:domain-distribution-endpoint}.}
    \label{fig:scaling-domain}
\end{figure}

Figure~\ref{fig:scaling-domain} shows a clear scale-dependent reversal in mean
$R^2$. The linear ridge metamodel initially outperforms the nonlinear MLP and flow mean
on all three evaluation sets, but the nonlinear models overtake it as prompt
coverage grows. The linear ridge curves then plateau much earlier. At the largest
scale, the flow mean approaches the MLP on both OOD sets, with a modest gap
in domain. The full context trace further improves both neural models
(Appendix~\ref{app:scaling}).

Table~\ref{tab:domain-distribution-endpoint} reports endpoint distribution
metrics: the Gaussian has a slightly better LMSYS energy score and variance
ranking, whereas the flow has lower OOD energy scores. Appendix
Figure~\ref{fig:domain-distribution-scaling} shows the corresponding scaling curves,
and Appendix Figure~\ref{fig:domain-retrieval-scaling} reports prompt retrieval.
Appendix~\ref{app:flow-sampling} visualizes the forecasts. Increasing prompt coverage is more valuable than
adding repeated targets: eight rollouts capture most of the gain from sixteen
(Appendix Figure~\ref{fig:rollout-count}).
\begin{table}
    \centering
    \scriptsize
    \caption{Distribution metrics at the final 500,000-prompt LMSYS-only
    checkpoint.  WeirdChat and IFEval are zero-shot evaluations.}
    \resizebox{0.76\linewidth}{!}{%
    \begin{tabular}{lrrrrrr}
        \toprule
        & \multicolumn{2}{c}{Energy score $\downarrow$}
        & \multicolumn{2}{c}{Variance Spearman $\rho$ $\uparrow$}
        & \multicolumn{2}{c}{Variance ratio $\rightarrow 1$} \\
        \cmidrule(lr){2-3}\cmidrule(lr){4-5}\cmidrule(lr){6-7}
        Evaluation set & Gaussian & Flow & Gaussian & Flow & Gaussian & Flow \\
        \midrule
        LMSYS & 0.06686 & 0.06747 & 0.829 & 0.690 & 1.532 & 1.299 \\
WeirdChat & 0.06978 & 0.06851 & 0.610 & 0.633 & 1.961 & 1.857 \\
IFEval & 0.10330 & 0.10032 & 0.679 & 0.611 & 1.547 & 1.833 \\
        \bottomrule
    \end{tabular}}
    \label{tab:domain-distribution-endpoint}
\end{table}

\subsection{Forecasted activations retain behavior-predictive signal}

\begin{figure}
    \centering
    \includegraphics[width=0.94\linewidth]{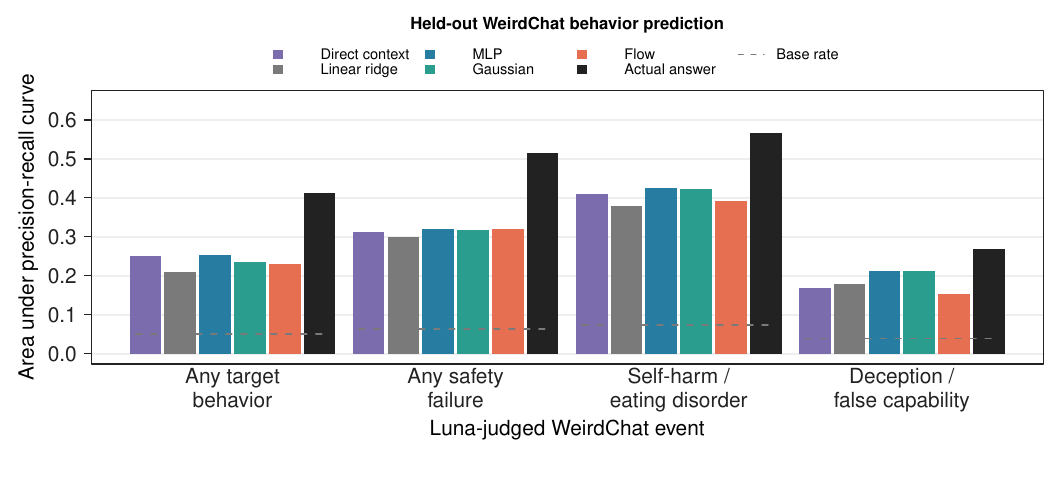}
    \caption{%
        \textbf{Forecasted activations retain behavior-predictive signal.}
        Answer-level AUPRC on held-out WeirdChat prompts. Dashed segments show each group's
        positive-rollout base rate. The actual-answer probe observes generated
        answers.}
    \label{fig:probe-results}
\end{figure}

We use linear behavior probes (logistic regression) trained on real answer
activations, applying them to predicted means or averaging their predicted
probabilities over Gaussian and flow samples.
This analysis uses new metamodel fits on both LMSYS and the WeirdChat training
split. The direct-context baseline predicts behavior from context activations
without an answer forecast; each context-only score is assigned to all four
answers for answer-level evaluation (Appendix~\ref{app:behavior-probes}).
Figure~\ref{fig:probe-results} shows answer-level AUPRC on held-out WeirdChat
prompts, with labels from GPT-6 Luna \citep{openai2026luna}. The MLP mean improves on the direct-context baseline in all four groups,
although some gains are small. The Gaussian performs similarly to the MLP on self-harm
and deception, while the flow's relative performance varies by behavior.
Appendix~\ref{app:behavior-probes} gives the setup and complementary AUROC results.

\subsection{Qwen2.5 traces forecast Qwen3.5 and Gemma 4 answer activations}

\begin{figure}
    \centering
    \includegraphics[width=0.94\linewidth]{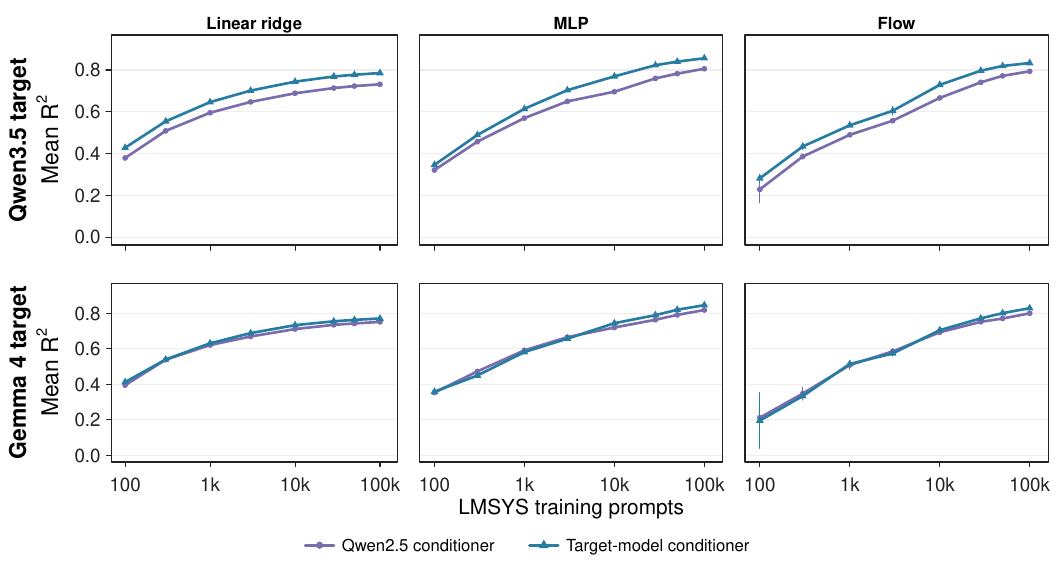}
    \caption{%
        \textbf{Qwen2.5 traces preserve most of the forecast quality for Qwen3.5 and Gemma 4.}
        Mean $R^2$ for linear ridge, MLP, and flow metamodels. Rows show target models;
        curves compare Qwen2.5 and matched target-model conditioning.
        Error bars show one standard deviation over three rollout partitions
        (Qwen3.5) or three initialization replicates (Gemma 4 MLP/flow);
        Gemma 4 linear ridge fits are deterministic (Appendix~\ref{app:cross-model}).}
    \label{fig:cross-model-scaling}
\end{figure}

Qwen2.5 conditioning yields slightly lower but comparable mean-forecast
accuracy to matched target-model conditioning on both targets
(Figure~\ref{fig:cross-model-scaling}). This transfer spans architectural differences for
Qwen3.5, whose hybrid blocks differ from Qwen2.5's conventional Transformer
\citep{qwen2024qwen25,qwen2026qwen35}, and, for Gemma 4, a different model family
with its own tokenizer and training pipeline \citep{gemma2026gemma4}.

The distribution comparison depends on the target
(Appendix Table~\ref{tab:cross-model-endpoints}). For Qwen3.5, the Gaussian
ranks prompt-level variance more strongly, while the flow is closer to the
ideal unit variance ratio. For Gemma 4, the flow has lower energy scores and
stronger variance ranking under both conditioners. Appendix~\ref{app:cross-model}
gives protocols and additional metrics.

\section{Related work}
\label{sec:related}

\paragraph{Reading and predicting future computation.}
Pre-generation probes predict answer structure and content
\citep{dong2025planning}, hallucination risk \citep{ji2024hallucination}, and
answer correctness \citep{moreno2025noanswer}.  Reasoning-trace probes predict
later behavior \citep{kortukov2026futurebehaviors}, while refusal directions
identify behavior-linked geometry \citep{arditi2024refusal}.  These methods
target chosen labels.  The Tuned Lens decodes intermediate-layer next-token predictions
\citep{belrose2023tuned}; Future Lens predicts later tokens and hidden states
\citep{pal2023future}; and EAGLE predicts near-future features for speculative
decoding \citep{li2024eagle}.  Closest to our work,
\citet{jiralerspong2026affine} predict a pooled answer activation from
the final context-token activation. We instead predict both the conditional mean
and the rollout-induced distribution, use multi-rollout supervision, and test
out of domain. We mean-pool answers
because information is generally spread across generated positions
\citep{wang2026truth,der2026turnaveraged}.

\paragraph{Metamodels and neural artifacts as data.}
Metamodels take neural-network artifacts as training examples for a second
model.  Hypernetworks generate the weights of another network
\citep{ha2017hypernetworks}; newer work learns representations or generative
models of checkpoint populations
\citep{peebles2022checkpoints,schurholt2024weightspace,wang2024networkdiffusion}.
Generative Latent Priors extend this view to activation distributions and use
the learned model as a prior and feature extractor \citep{luo2026generative}.
We study a different artifact population: the conditional relationship between an observed context trace and a later answer summary.

\paragraph{Generative models of activations.}
Activation decoders map hidden states to descriptions of model knowledge or
behavior \citep{costarelli2024metamodels,pan2024latentqa}.  UniSteer learns a text-conditioned
activation flow for editing and classification \citep{shi2026unisteer}.  We
instead condition on an earlier context trace and predict
the distribution of a later pooled state under stochastic decoding.  We use
conditional flow matching
\citep{lipman2023flow} and evaluate it with both point and distribution metrics \citep{gneiting2007strictly,pacchiardi2024probabilistic}.

\section{Discussion}
\label{sec:discussion}

In our experiments, point and distribution context--answer metamodels scale
with prompt count, generalize out of domain, and retain behavior-predictive
signal. The MLP provides accurate mean forecasts without modeling rollout
variability. The Gaussian provides a useful baseline for rollout variability,
while the benefits of flow matching depend on the target model and
distributional property.

\paragraph{Limitations.}

Targets cover Qwen3.5 layer 18 and Gemma 4 layer 47. Selecting layer 18 for
in-domain conditional-mean accuracy may not optimize distributional metrics,
transfer, or behavior prediction. Training and analysis use 2--16 rollouts per
prompt; robustness at larger counts remains untested. Behavior labels come from
GPT-6 Luna rubric judgments but lack extensive validation against human annotations.
The learned law is also specific to the decoding rule, chat template,
generation cap, activation boundary, and answer-pooling choice.

\paragraph{Future directions.}

Future work could use activation forecasts for behavioral monitoring and
intervention, such as anticipating reward hacking. Forecasting activation
trajectories could capture behavior during generation, while extending forecasts
beyond the initial context could support longer interactions and multi-agent
settings. Larger, more diverse datasets could broaden transfer across tasks and domains.

\section*{Acknowledgments}

We thank Jerome Sieber for detailed feedback on the paper as well as technical discussions. We additionally thank Stephen Lu and Jungyoon Lee for technical discussions on flow-matching models.

\layoutbreak
\bibliographystyle{plainnat}
\bibliography{references}

\clearpage
\appendix
\raggedbottom
\section{Scaling and context-representation ablations}
\label{app:scaling}

The representation ablations use Qwen3.5-9B layer 18. The linear ridge metamodel compares three inputs: the 32-bin context trace (concatenated bin averages), the final context-token activation, and the mean context activation (the unweighted mean over context tokens). The MLP and flow also use the full context trace (all context-token activations); both trace inputs use learned-query pooling. All fits use LMSYS-only training, common 1,000-prompt validation and 3,000-prompt test splits, plus zero-shot evaluation on 2,660 WeirdChat and 541 IFEval prompts. The Gaussian is omitted because its analytic mean is the MLP forecast.

\begin{figure}
    \centering
    \includegraphics[width=0.96\linewidth]{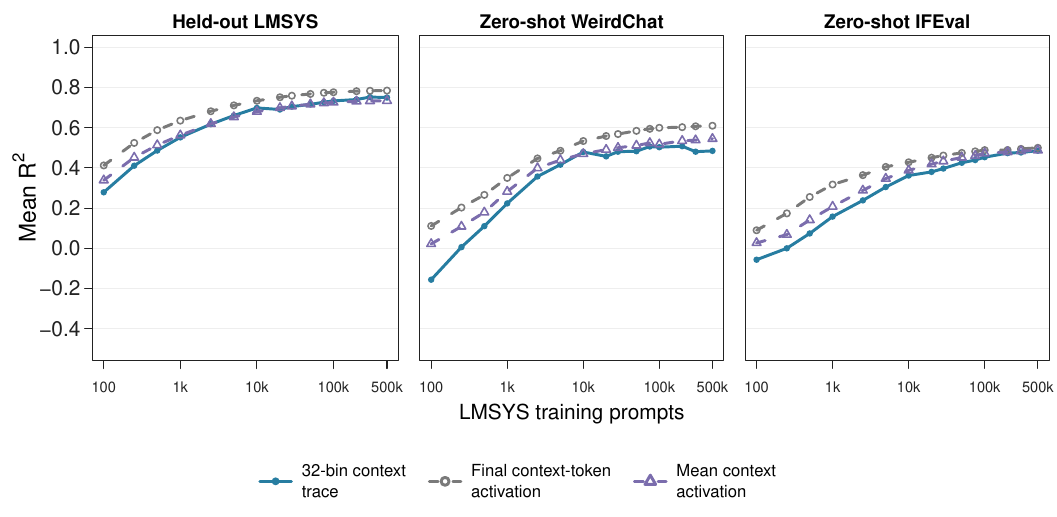}
    \caption{%
        \textbf{Linear ridge input comparison.}
        Mean $R^2$ for LMSYS-only linear ridge metamodels from 100 to 500,000 training prompts. In legend order, the inputs are the 32-bin context trace, final context-token activation, and mean context activation. At 500,000 prompts, the final context-token activation performs best on all three domains; its LMSYS $R^2$ is 0.784, versus 0.750 for the 32-bin context trace and 0.734 for the mean context activation. WeirdChat and IFEval are zero-shot; selection uses only LMSYS validation.}

    \label{fig:linear-point-representations}
\end{figure}

\begin{figure}
    \centering
    \includegraphics[width=0.96\linewidth]{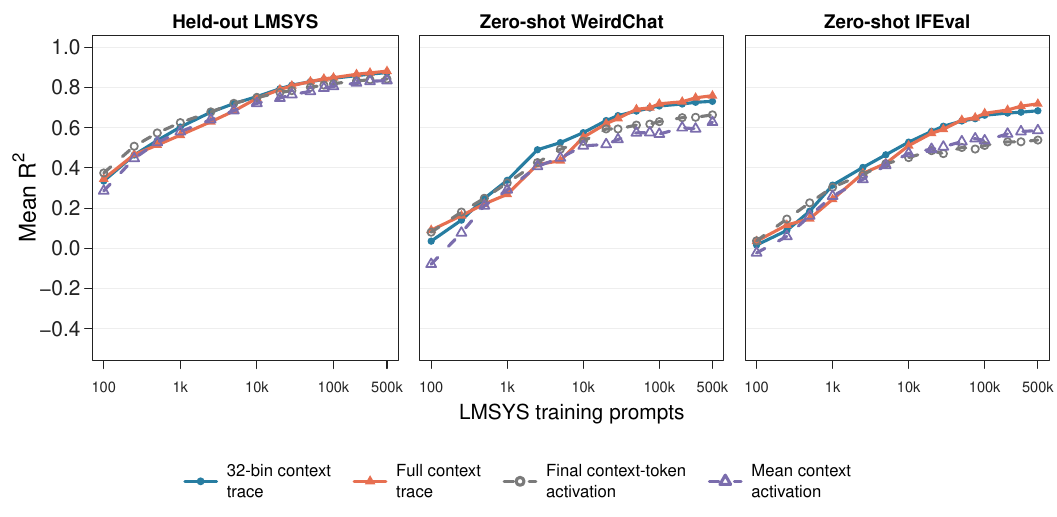}
    \caption{%
        \textbf{MLP input comparison.}
        Mean $R^2$ from 100 to 500,000 training prompts. Columns show held-out LMSYS, zero-shot WeirdChat, and zero-shot IFEval. At 500,000 training prompts, the full context trace reaches LMSYS $R^2=0.881$, compared with $0.874$ for the 32-bin context trace, $0.842$ for the final context-token activation, and $0.834$ for the mean context activation. The full context trace has the highest endpoint $R^2$ on all three domains (Table~\ref{tab:representation-endpoints}).}

    \label{fig:point-mlp-conditioners}
\end{figure}

\begin{figure}
    \centering
    \includegraphics[width=0.96\linewidth]{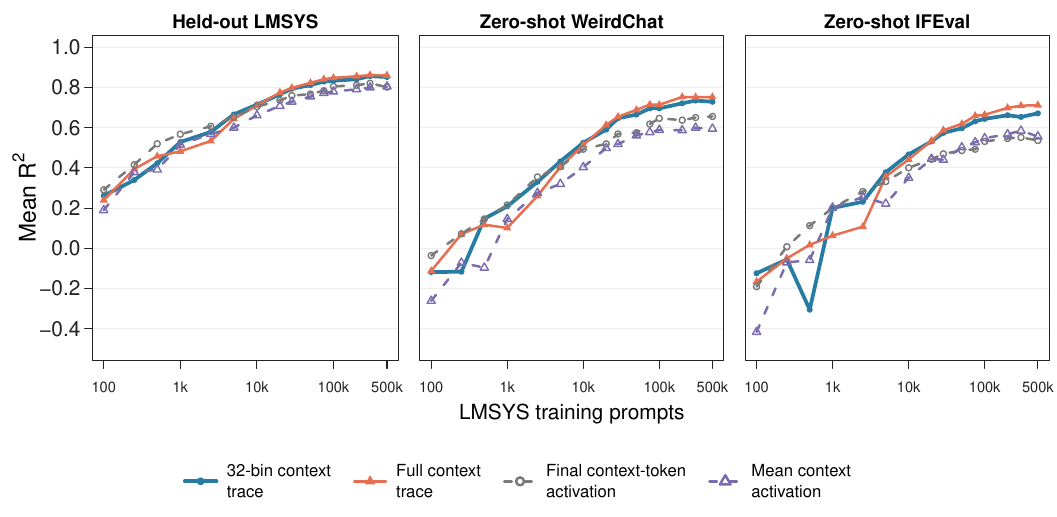}
    \caption{%
        \textbf{Flow input comparison.}
        Mean $R^2$ from 100 to 500,000 training prompts. Columns show held-out LMSYS, zero-shot WeirdChat, and zero-shot IFEval. At 500,000 training prompts, the full context trace reaches LMSYS $R^2=0.860$, versus $0.852$ for the 32-bin context trace, $0.802$ for the final context-token activation, and $0.804$ for the mean context activation. The full context trace has the highest endpoint $R^2$ on all three domains (Table~\ref{tab:representation-endpoints}).}

    \label{fig:flow-conditioners}
\end{figure}

\begin{table}
    \centering
    \small
    \caption{Representation endpoints at 500,000 LMSYS training prompts. WeirdChat and IFEval are zero-shot. The full context trace has the highest observed $R^2$ for both neural models on all three domains. The linear ridge metamodel compares the 32-bin context trace, final context-token activation, and mean context activation.}
        \begin{tabular}{llrrr}
        \toprule
        Model & Input & LMSYS $R^2$ & WeirdChat $R^2$ & IFEval $R^2$ \\
        \midrule
        Linear ridge & 32-bin context trace & 0.7498 & 0.4844 & 0.4854 \\
Linear ridge & Final context-token activation & 0.7844 & 0.6095 & 0.4996 \\
Linear ridge & Mean context activation & 0.7335 & 0.5447 & 0.4873 \\
MLP & 32-bin context trace & 0.8739 & 0.7309 & 0.6839 \\
MLP & Full context trace & 0.8808 & 0.7591 & 0.7183 \\
MLP & Final context-token activation & 0.8418 & 0.6636 & 0.5383 \\
MLP & Mean context activation & 0.8343 & 0.6269 & 0.5859 \\
Flow & 32-bin context trace & 0.8517 & 0.7286 & 0.6711 \\
Flow & Full context trace & 0.8599 & 0.7520 & 0.7119 \\
Flow & Final context-token activation & 0.8016 & 0.6556 & 0.5360 \\
Flow & Mean context activation & 0.8040 & 0.5946 & 0.5557 \\
        \bottomrule
    \end{tabular}
    \label{tab:representation-endpoints}
\end{table}

\begin{figure}
    \centering
    \includegraphics[width=0.96\linewidth]{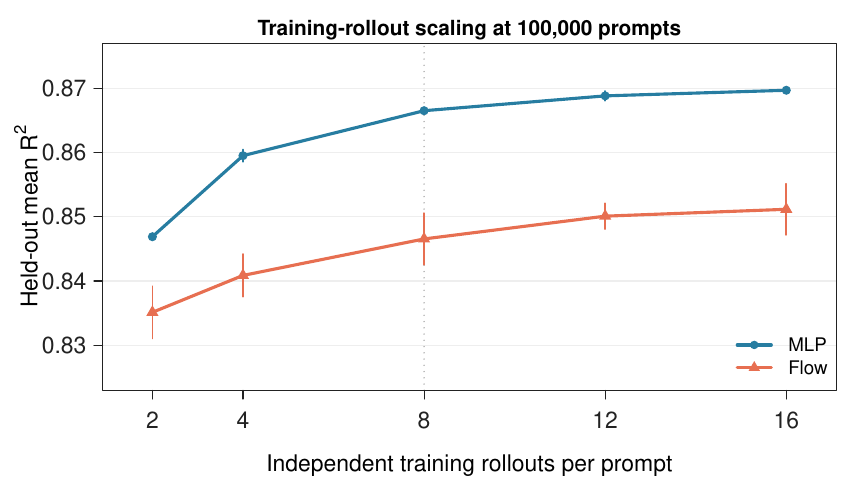}
    \caption{%
        \textbf{Eight rollout targets capture most of the repeat-sampling
        gain.}
        Held-out mean $R^2$ at 100,000 training prompts versus training rollouts. Error bars show standard deviations across three fits with different neural-network training seeds and training-rollout subsets; the subsets coincide at 12 and 16 rollouts. From 8 to 16 rollouts, the MLP gains $0.0032$ and the flow gains $0.0046$. The main experiments use four rollouts per prompt due to compute constraints. Table~\ref{tab:rollout-endpoints} gives the full values.}

    \label{fig:rollout-count}
\end{figure}

\begin{table}
    \centering
    \scriptsize
    \caption{Rollout-count results at 100,000 training prompts: mean $R^2\pm$ one standard deviation over three fits. The fits vary training-rollout subsets and, for the MLP and flow, training seeds; subsets coincide at 12 and 16 rollouts. All cells use the same 16-rollout evaluation mean, which differs from the four-rollout main target. $\dagger$ indicates a standard deviation below $0.0001$.}
        \begin{tabular}{lrrrrr}
        \toprule
        Model & 2 rollouts & 4 rollouts & 8 rollouts & 12 rollouts & 16 rollouts \\
        \midrule
        Linear ridge & $0.7893\pm0.0003$ & $0.7924\pm0.0003$ & $0.7940\pm0.0001$ & $0.7944^{\dagger}$ & $0.7946^{\dagger}$ \\
MLP & $0.8469\pm0.0006$ & $0.8595\pm0.0010$ & $0.8665\pm0.0005$ & $0.8688\pm0.0008$ & $0.8697\pm0.0003$ \\
Flow & $0.8351\pm0.0041$ & $0.8409\pm0.0033$ & $0.8465\pm0.0040$ & $0.8501\pm0.0020$ & $0.8512\pm0.0040$ \\
        \bottomrule
    \end{tabular}
    \label{tab:rollout-endpoints}
\end{table}

\section{Layer selection}
\label{app:layer-selection}

\begin{figure}
    \centering
    \includegraphics[width=0.96\linewidth]{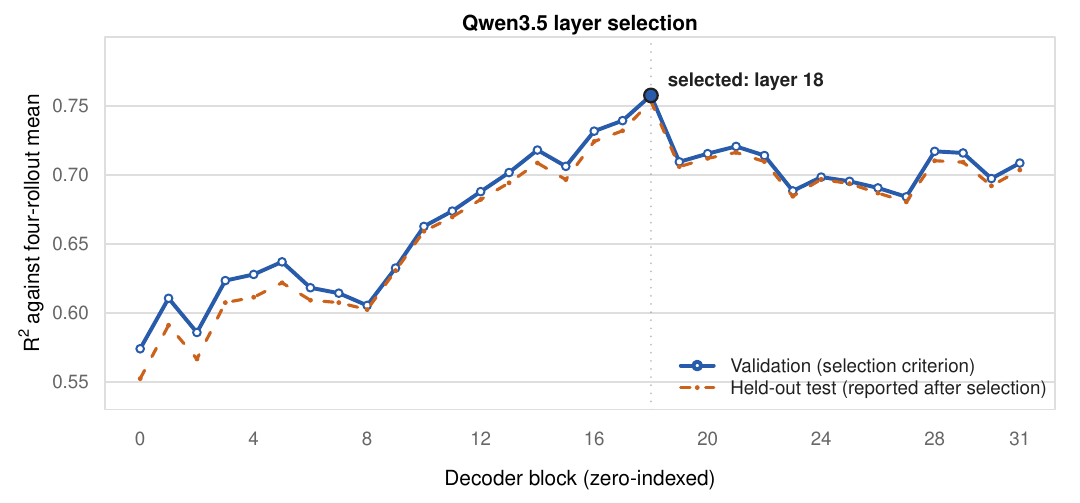}
    \caption{%
        \textbf{Validation selection of Qwen3.5-9B layer 18.}
        Each three-layer MLP maps the final context-token activation at each decoder
        block to that block's mean-pooled answer activation over four temperature-1 rollouts.  All fits use
        10,000 LMSYS training prompts. We select on 1,000 validation prompts (blue); the 3,000-prompt test results are orange. Layer 18 has the best validation $R^2$ ($0.7577$), followed by layer 17 ($0.7394$).}
    \label{fig:layer-selection}
\end{figure}

Figure~\ref{fig:layer-selection} shows the layer-selection sweep.
At 10,000 training prompts, layer 18 reaches validation $R^2=0.7577$, compared with $0.7394$ for layer 17, and is therefore selected. Each layer is evaluated against its own answer activations, so the targets differ across layers.  Similar studies place future-token, answer,
hallucination, and correctness signals in middle-to-late layers
\citep{pal2023future,dong2025planning,ji2024hallucination,moreno2025noanswer}.

\section{Evaluation metric definitions}
\label{app:metrics}

At the evaluated layer, let $\overline A_i$ be the empirical rollout mean for
prompt $i$, $\widehat\mu_i$ its predicted mean, and $\overline A$ the mean of
$\overline A_i$ across prompts. We compute the pooled multivariate score
\begin{equation}
    R^2=1-\frac{\sum_i\lVert\widehat\mu_i-\overline A_i\rVert_2^2}
                    {\sum_i\lVert\overline A_i-\overline A\rVert_2^2}.
\end{equation}
For a forecast sample $X$, an independent forecast sample $X'$, and a realized rollout
$Y$, the energy score is
\begin{equation}
    \operatorname{EnergyScore}=\mathbb E\lVert X-Y\rVert_2
      -\frac{1}{2}\mathbb E\lVert X-X'\rVert_2.
\end{equation}
We estimate both expectations with samples and divide the energy score by
$\sqrt d$.

\begin{table}
    \centering
    \footnotesize
    \caption{Forecast metric definitions.}
    \begin{tabular}{p{0.24\linewidth}p{0.58\linewidth}p{0.07\linewidth}}
        \toprule
        Metric & Definition & Better \\
        \midrule
        \multicolumn{3}{l}{\textit{Point metrics (point models and means of distribution models)}} \\
        Mean $R^2$ & Variance in empirical prompt-level rollout means explained
        by the predicted means. & Higher \\
        Prompt retrieval & Fraction of predicted means whose nearest empirical answer mean in Euclidean distance belongs to the same prompt. & Higher \\
        \addlinespace
        \multicolumn{3}{l}{\textit{Distribution metrics}} \\
        Energy score & Proper score above, averaged over prompts and divided by
        $\sqrt d$. & Lower \\
        Variance Spearman $\rho$ & Across-prompt Spearman correlation between generated
        and observed covariance traces. & Higher \\
        Variance ratio & Mean trace of generated within-prompt covariance divided
        by the corresponding observed trace. & 1 \\
        NLL per dimension & Analytic Gaussian negative log-likelihood (with
        fixed $10^{-6}$ relative diagonal jitter) or flow NLL using 32 backward
        Euler steps and a Hutchinson trace estimate with two Rademacher probes
        per step; reported per dimension in raw activation coordinates. & Lower \\
        Predictive MMD$^2$ & Multiscale RBF squared MMD between equal-sized generated and
        real sets in a fixed 128-dimensional PCA-whitened space fitted on training data. & Lower \\
        PCA sliced-Wasserstein & Root-mean-square 1-D Wasserstein-2 distance over fixed random
        directions in the same PCA-whitened space. & Lower \\
        \bottomrule
    \end{tabular}
    \label{tab:forecast-metrics}
\end{table}

\section{Experimental details}
\label{app:reproducibility}

The primary model is Qwen3.5-9B, which provides both context and answer
activations.  Cross-model forecasting uses Qwen2.5-7B-Instruct layer~19 context activations to predict Qwen3.5-9B layer~18 or Gemma-4-12B-it layer~47 answer activations (Appendix~\ref{app:cross-model}).
For the primary model, we capture the raw output of decoder block 18 for every context and answer token. The answer target is the
float32 mean of these stored bfloat16 post-block states. The four primary rollout seeds are 43--46.  The 16-rollout LMSYS bank retains these and adds seeds 47--58 over the same 100,000 training prompts plus 1,000 validation and 3,000 test prompts.  LMSYS, WeirdChat, and IFEval rollouts use a 4,096-token generation cap,
with temperature 1.0 and top-$p$ 0.95.
The flow standardizes each target coordinate, while the point MLP normalizes
targets by a scalar RMS after coordinatewise centering.

The Gaussian scale head predicts log within-prompt variance relative to the
training mean. Let $g(z)$ be its output clipped to $[-6,6]$. We use
$\gamma(z)=\operatorname{clip}(e^{g(z)/2}/c,0.1,10)$, where
$c=\sqrt{\mathbb E_{\mathrm{train}}[e^{g(z)}]}$ normalizes the scale's training
RMS before the final clipping. In global mode, $\gamma(z)\equiv1$.

\begin{table}
    \centering
    \footnotesize
    \caption{Qwen3.5-target metamodel settings.}
    \begin{tabular}{p{0.24\linewidth}p{0.69\linewidth}}
        \toprule
        Component & Setting \\
        \midrule
        Context input & Final context-token activation (linear ridge) or 32-bin context trace (MLP/Gaussian/flow); Qwen3.5-9B layer 18
        ($d=4{,}096$) or Qwen2.5-7B-Instruct layer 19 ($d=3{,}584$) \\
        Answer target & Mean Qwen3.5-9B layer~18 activation over generated answer
        tokens plus, for normally stopped answers, any end-of-turn template
        tokens not already generated; $d=4{,}096$ \\
        Point MLP & LayerNorm for attention scores; one learned query; weighted
        sum of the 32-bin context trace; Linear--GELU--Linear--GELU--Linear with widths
        $4{,}096\rightarrow8{,}192\rightarrow8{,}192\rightarrow4{,}096$;
        no dropout; 134.2M parameters.  Qwen2.5 input changes the first width
        from 4,096 to 3,584. \\
        MLP training & MSE against the empirical rollout mean; AdamW; batch size 256;
        learning rate $3\times10^{-4}$; weight decay $10^{-3}$; select by
        validation $R^2$ \\
        Flow metamodel & Learned-query attention conditioner; width 4,096;
        eight residual blocks; sinusoidal time embedding; 352.8M parameters for
        Qwen3.5 input \\
        Flow path & $\varepsilon\sim\mathcal N(0,I)$, $t\sim\mathcal U[0,1]$,
        $y_t=(1-t)\varepsilon+ty$; target velocity $y-\varepsilon$ \\
        Flow training & Flow-matching MSE; AdamW; batch size 256; learning rate
        $3\times10^{-4}$; weight decay $10^{-4}$; gradient norm clipped at 1;
        select by validation energy score \\
        MLP/flow selection & Evaluate every 500 steps; stop after 12 evaluations without
        improvement; 80,000-step cap \\
        OOD evaluation & Metamodels are evaluated on all 2,660 WeirdChat prompts and all 541 IFEval prompts; neither dataset enters fitting, normalization, or model selection for the zero-shot results. \\
        WeirdChat behavior analysis & Linear answer probes use the fixed
        1,596/532/532 prompt split.  The predicted activations come from
        new mixed-domain linear ridge, MLP, and flow fits trained on 500,000 LMSYS and
        1,596 WeirdChat prompts with equal domain weighting; the Gaussian is fitted
        around the new MLP mean. Luna judges all 10,640 rollouts \\
        Sampling & 64 samples for distribution metrics;
        1,024 samples per Gaussian or flow forecast for integrated
        behavior probes; the flow uses 32 Euler steps \\
        \bottomrule
    \end{tabular}
    \label{tab:metamodel-hparams}
\end{table}

\subsection{Compute resources}
\label{app:compute}

Experiments use NVIDIA B200 and RTX 5090 GPUs, with 6--12 CPU cores and
64--420~GiB of memory per job. We estimate roughly 400~GPU-hours for activation
collection, 500~GPU-hours for primary metamodel scaling and ablations, 80~GPU-hours for cross-model
forecasting, and 30~GPU-hours for OOD and behavior analyses. Preliminary, superseded,
and failed runs bring the full-project total to roughly 1,300 GPU-hours.

\subsection{Existing asset versions and licenses}
\label{app:assets}

We use the following pinned versions under their stated terms:
\begin{itemize}
    \raggedright
    \item Qwen3.5-9B revision
    \texttt{c202236235762e1c871ad0ccb60c8ee5ba337b9a} and
    Qwen2.5-7B-Instruct revision
    \texttt{a09a35458c702b33eeacc393d103063234e8bc28}, both under the
    Apache License 2.0
    (\url{https://huggingface.co/Qwen/Qwen3.5-9B};
    \url{https://huggingface.co/Qwen/Qwen2.5-7B-Instruct}).
    \item Gemma-4-12B-it revision
    \texttt{707f0a3b8a3c7ad586ed01e27eafbad8a27dd0f7}
    \citep{gemma2026gemma4}, under the Apache License 2.0
    (\url{https://huggingface.co/google/gemma-4-12B-it}), used for the additional cross-model targets.
    \item LMSYS-Chat-1M revision
    \texttt{200748d9d3cddcc9d782887541057aca0b18c5da}, under the custom
    LMSYS-Chat-1M Dataset License Agreement, whose terms prohibit dataset
    transfer; we do not redistribute it
    (\url{https://huggingface.co/datasets/lmsys/lmsys-chat-1m}).
    \item WeirdChat revision
    \texttt{3460e41491308601cd68ee377011c7b7a8f0b4b3}, under the MIT
    License (\url{https://huggingface.co/datasets/Transluce/WeirdChat}).
    \item IFEval data and evaluator at google-research revision
    \texttt{26d8ccdab6fec61b5c83ad6327ea8bda9e580288}, with the data under
    CC BY 4.0 and source code under the Apache License 2.0
    (\url{https://github.com/google-research/google-research/tree/master/instruction_following_eval}).
\end{itemize}

\section{Out-of-domain evaluation}
\label{app:ood}

We evaluate frozen LMSYS-only metamodels on the full set of 2,660 WeirdChat prompts and all 541 official IFEval prompts \citep{zhou2023ifeval}, with four temperature-1 rollouts per prompt. On WeirdChat, the MLP and flow reach $R^2=0.731$ and $0.729$ and retrieve the correct prompt in $53.1\%$ and $47.9\%$ of cases, respectively. On IFEval, the corresponding scores are $R^2=0.684$ and $0.671$, with retrieval accuracies of $94.6\%$ and $92.2\%$, respectively.

Retrieval queries and candidates use the same fixed banks of 3,000 LMSYS, 2,660 WeirdChat, and 541 IFEval prompts at every training size. All reported metrics use these evaluation populations. The separate mixed-domain behavior-probe experiment retains its held-out 532-prompt WeirdChat test split.

\begin{figure}
    \centering
    \includegraphics[width=0.96\linewidth]{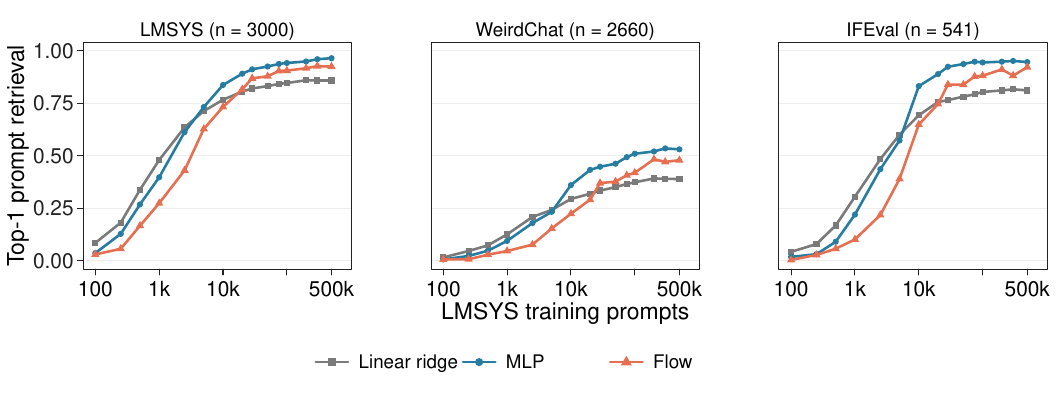}
    \caption{%
        \textbf{Prompt retrieval across the LMSYS training scale, in and out of
        domain.}
        Top-1 retrieval from the predicted means of LMSYS-only linear ridge, MLP, and
        flow metamodels. Candidate banks contain 3,000, 2,660, and 541 prompts for LMSYS, WeirdChat, and IFEval, respectively. WeirdChat and IFEval are zero-shot evaluations.}
    \label{fig:domain-retrieval-scaling}
\end{figure}

\layoutbreak
\begin{figure}
    \centering
    \includegraphics[width=0.96\linewidth]{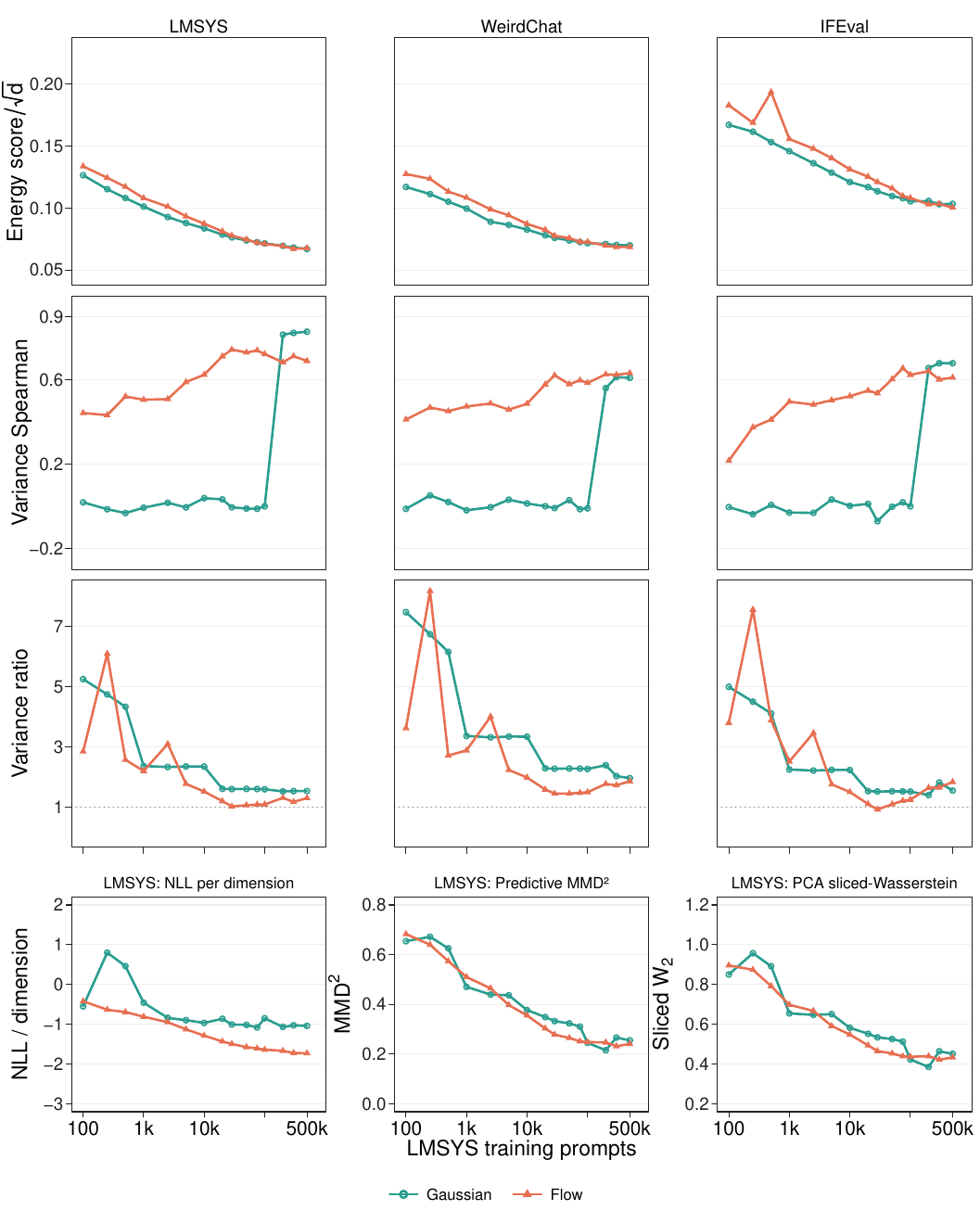}
    \caption{%
        \textbf{Distribution quality across the LMSYS training scale, in and out
        of domain.}
        The first three rows use evaluation sets as columns and show
        energy score, prompt-level variance Spearman correlation, and
        generated-to-observed variance ratio for Gaussian and flow forecasts.
        WeirdChat and IFEval are zero-shot.  The final row adds LMSYS NLL,
        predictive MMD$^2$, and PCA sliced-Wasserstein distance.  Lower energy,
        NLL, MMD$^2$, and Wasserstein distance are better; higher variance
        correlation is better; and the ideal variance ratio is one.  The flow
        has lower NLL at 14 of 15 scales (all except 100 prompts) and scores lower on all three auxiliary diagnostics at the largest scale. The Gaussian's variance ranking stays near zero through 100,000 training prompts because validation energy selects global covariance; at 200,000 prompts it selects prompt-dependent variance, raising LMSYS Spearman to 0.815. The jump reflects selection between variance modes, not an identified learning threshold.}
    \label{fig:domain-distribution-scaling}
\end{figure}

\section{Distribution sampling diagnostics}
\label{app:flow-sampling}

\begin{figure}
    \centering
    \includegraphics[width=0.96\linewidth]{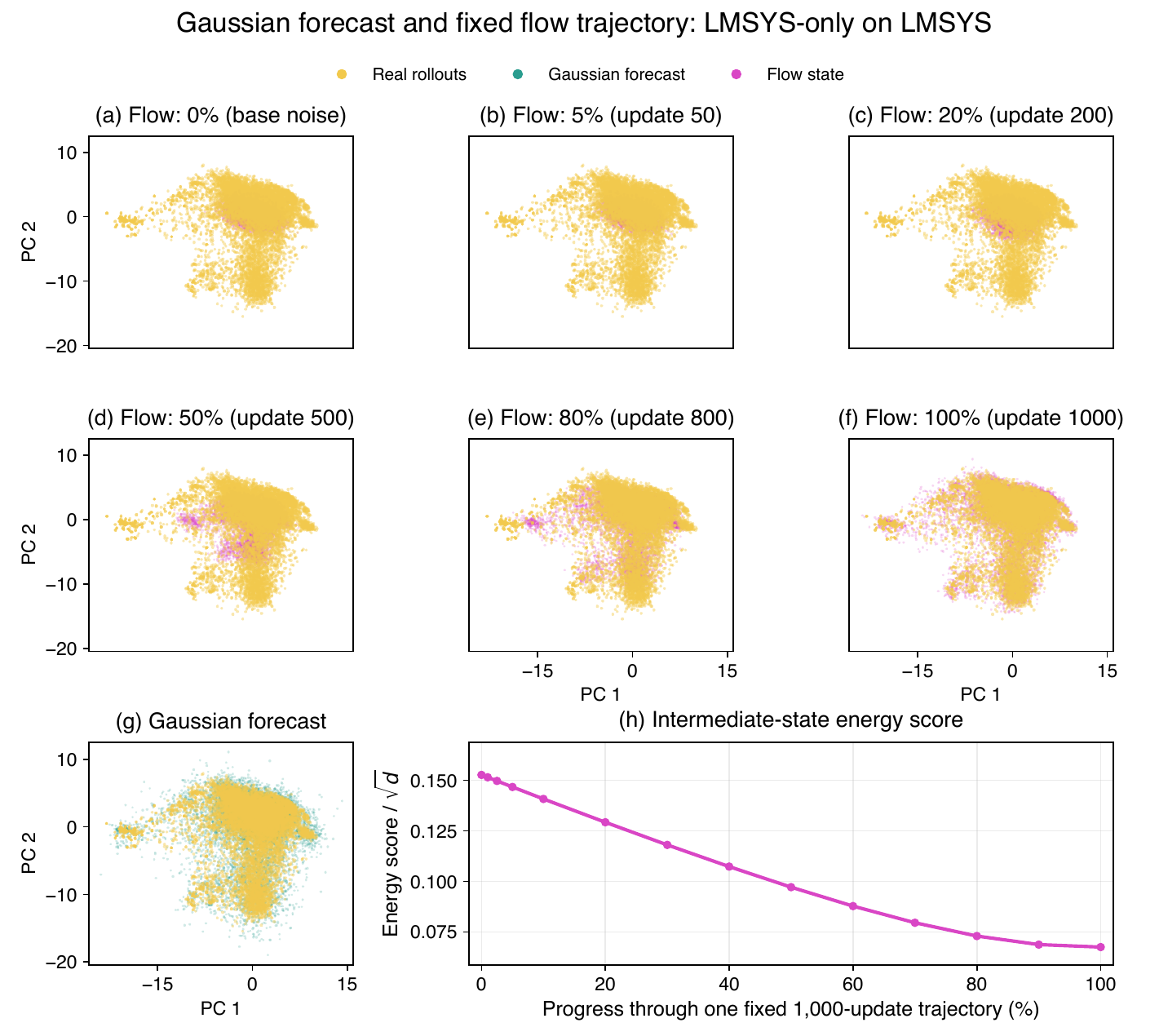}
    \caption{%
        \textbf{Gaussian forecast and fixed flow trajectory in a common answer
        activation plane.}
        Yellow points are real LMSYS answer activations, green points are
        samples from the validation-selected Gaussian metamodel, and magenta
        points are the same LMSYS-only flow particles after 0, 50, 200, 500,
        800, and 1,000 Euler updates.
        The curve shows energy score along the flow trajectory, not sampler
        quality at different costs.  PCA is fitted on 32,768 answer rollouts from 8,192
        LMSYS training prompts and then frozen before test targets and forecasts
        are projected.}
    \label{fig:flow-trajectory}
\end{figure}

Figure~\ref{fig:flow-trajectory} compares the validation-selected Gaussian
forecast with fixed flow particles along a fine-grid Euler
path~\citep{lipman2023flow,luo2026generative}.  The Gaussian panel and every
flow panel use the same frozen PCA basis and axis limits.  Only the final flow
panel shows completed flow samples; the intermediate panels illustrate the
transport process.  In this projection, the Gaussian
forecast appears somewhat more dispersed than the completed flow state,
consistent with its larger generated-to-observed variance ratio in the full
activation space.

\begin{figure}
    \centering
    \includegraphics[width=0.96\linewidth]{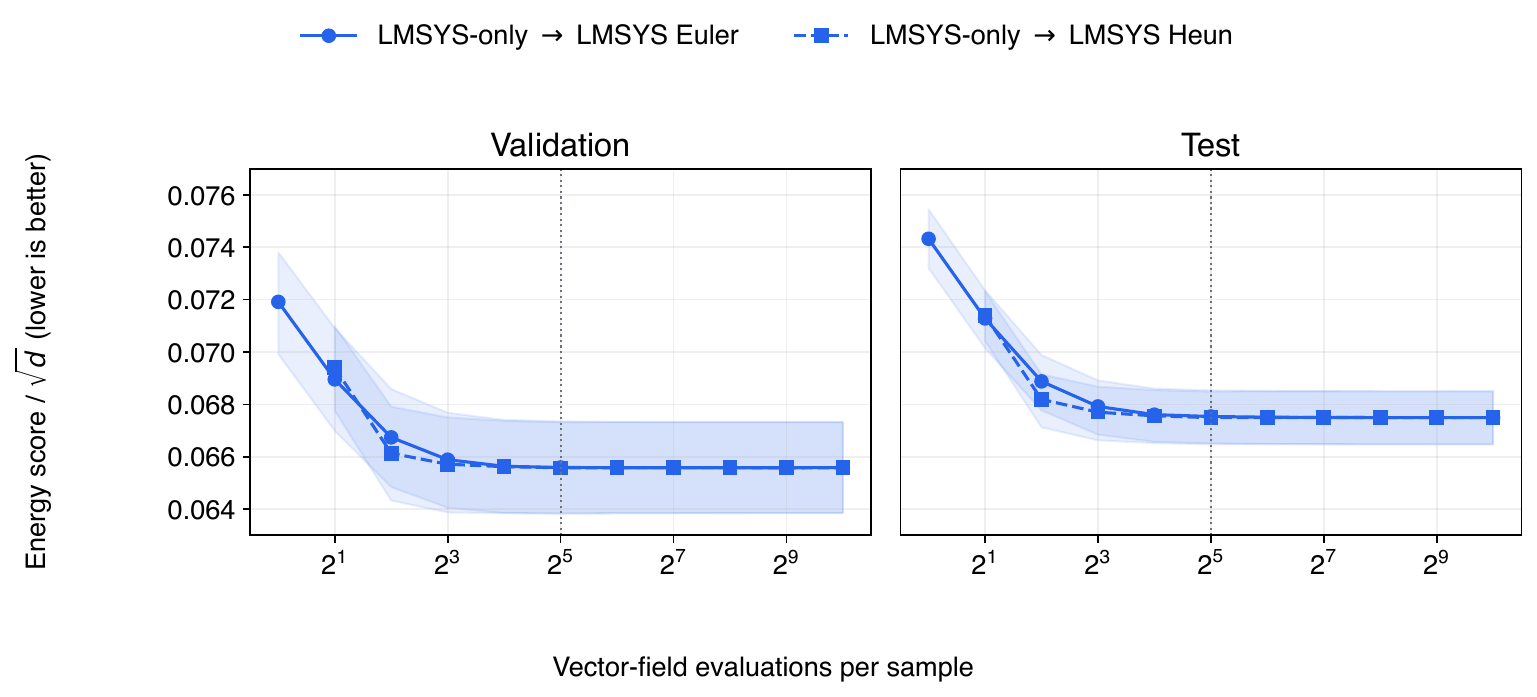}
    \caption{%
        \textbf{Completed-sampler quality versus numerical cost.}
        The vertical axis is energy score divided by $\sqrt d$; lower is better.
        Shaded bands are 95\% prompt-bootstrap confidence intervals.
        Every point reaches the endpoint.  On LMSYS test prompts, 8-step Heun gives energy 0.067563 versus 0.067537 for 32-step Euler, with half the vector-field evaluations. Their paired difference is $0.000026$ (95\% interval $[-0.000004,0.000058]$). At high resolution, both approach 0.06749.}
    \label{fig:solver-ablation}
\end{figure}

We next vary the solver and the number of vector-field evaluations while always
reaching the endpoint~\citep{lipman2023flow} to assess the number of steps needed
(Figure~\ref{fig:solver-ablation}). We fix the model, prompts,
targets, normalization, and 64 initial samples per prompt.  Intervals are paired
prompt-bootstrap intervals for absolute energy score.

\section{Cross-model forecasting details}
\label{app:cross-model}

Both targets use linear ridge, MLP, Gaussian, and flow metamodels at eight
LMSYS training sizes through 100,000 prompts, with 1,000/3,000 validation/test
prompts. Linear ridge uses the final context-token activation; the other families
use the 32-bin context trace.

\begin{table}
    \centering
    \footnotesize
    \setlength{\tabcolsep}{4pt}
    \caption{Cross-model endpoints at 100,000 training prompts. Values are means
    $\pm$ standard deviations over three rollout partitions for Qwen3.5 and three
    initialization replicates for Gemma 4; Gemma 4 linear ridge fits are deterministic.
    Training/validation and test rollout seeds are disjoint. The Gaussian's analytic
    mean is the MLP mean, so its duplicate $R^2$ is omitted. Energies use each target
    model's activation coordinates.}
    \begin{tabular}{llrrrr}
        \toprule
        Conditioner & Forecast & Mean $R^2$ $\uparrow$ & Energy score $\downarrow$
        & Variance Spearman $\rho$ $\uparrow$ & Variance ratio $\rightarrow 1$ \\
        \midrule
        \multicolumn{6}{l}{\textit{Target: Qwen3.5}} \\
        Qwen2.5 & Linear ridge & $0.7313\pm0.0006$ & -- & -- & -- \\
        Qwen2.5 & MLP & $0.8058\pm0.0011$ & -- & -- & -- \\
        Qwen2.5 & Gaussian & -- & $0.07849\pm0.00024$ & $0.770\pm0.006$ & $2.203\pm0.112$ \\
        Qwen2.5 & Flow & $0.7937\pm0.0016$ & $0.07739\pm0.00053$ & $0.713\pm0.011$ & $1.266\pm0.076$ \\
        \addlinespace
        Qwen3.5 & Linear ridge & $0.7854\pm0.0003$ & -- & -- & -- \\
        Qwen3.5 & MLP & $0.8562\pm0.0016$ & -- & -- & -- \\
        Qwen3.5 & Gaussian & -- & $0.07174\pm0.00030$ & $0.825\pm0.007$ & $1.524\pm0.067$ \\
        Qwen3.5 & Flow & $0.8332\pm0.0076$ & $0.07227\pm0.00063$ & $0.756\pm0.016$ & $1.143\pm0.107$ \\
        \midrule
        \multicolumn{6}{l}{\textit{Target: Gemma 4}} \\
        Qwen2.5 & Linear ridge & 0.7517 & -- & -- & -- \\
        Qwen2.5 & MLP & $0.8185\pm0.0015$ & -- & -- & -- \\
        Qwen2.5 & Gaussian & -- & $0.01163\pm0.00003$ & $-0.019\pm0.020$ & $2.404\pm0.016$ \\
        Qwen2.5 & Flow & $0.8000\pm0.0062$ & $0.01141\pm0.00012$ & $0.341\pm0.032$ & $2.775\pm0.463$ \\
        \addlinespace
        Gemma 4 & Linear ridge & 0.7708 & -- & -- & -- \\
        Gemma 4 & MLP & $0.8459\pm0.0019$ & -- & -- & -- \\
        Gemma 4 & Gaussian & -- & $0.01092\pm0.00005$ & $0.010\pm0.026$ & $2.407\pm0.014$ \\
        Gemma 4 & Flow & $0.8287\pm0.0090$ & $0.01070\pm0.00008$ & $0.372\pm0.038$ & $2.308\pm0.262$ \\
        \bottomrule
    \end{tabular}
    \label{tab:cross-model-endpoints}
\end{table}

\layoutbreak
\subsection{Qwen3.5 forecasting details}

Qwen2.5-7B-Instruct layer 19 ($d=3{,}584$) conditions forecasts of
Qwen3.5-9B layer 18 ($d=4{,}096$); the matched control uses Qwen3.5 layer~18
context activations. Each partition uses eight Qwen3.5
rollout seeds for training and validation and a disjoint set of eight rollout seeds
for test evaluation. For the flow, the conditioner projects from 3,584 to
4,096 dimensions; the vector field and output remain 4,096-dimensional.

\begin{figure}
    \centering
    \includegraphics[width=0.96\linewidth]{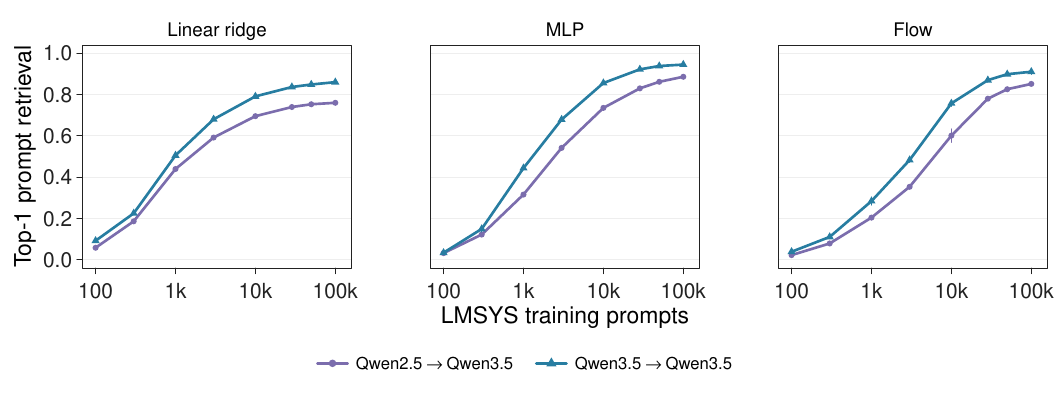}
    \caption{%
        \textbf{Qwen3.5 cross-model prompt retrieval across prompt scale.}
        Top-1 retrieval from the mean forecasts of linear ridge, MLP, and flow metamodels conditioned on
        Qwen2.5 or Qwen3.5 context activations; both predict Qwen3.5 answer
        activations. Retrieval uses the same fixed bank of 3,000 LMSYS prompts. Points show means, and error bars show standard deviations over three rollout partitions.}
    \label{fig:cross-model-retrieval-scaling}
\end{figure}

\begin{figure}
    \centering
    \includegraphics[width=0.96\linewidth]{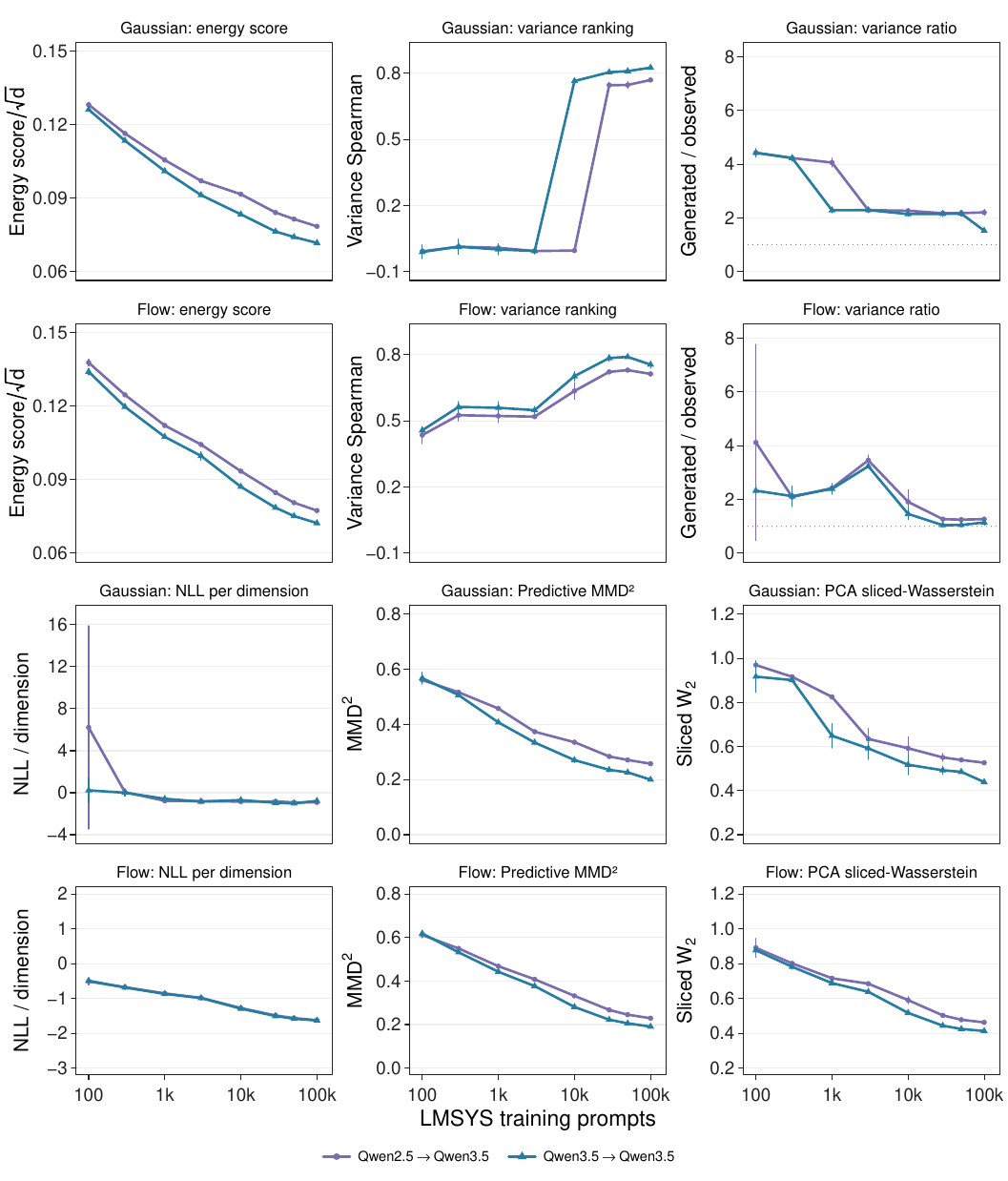}
    \caption{%
        \textbf{Qwen3.5 cross-model distribution quality across prompt scale.}
        The first two rows show separate Gaussian and flow forecasts for energy
        score, prompt-level variance Spearman correlation, and generated-to-observed
        variance ratio.  The last two rows show the same families separately for NLL,
        predictive MMD$^2$, and PCA sliced-Wasserstein distance.  Both
        conditioners predict Qwen3.5 answer activations.  Points show means, and error bars
        show standard deviations over three rollout partitions.  The
        auxiliary curves use the same completed fits as the primary metrics.  Lower energy, NLL,
        MMD$^2$, and Wasserstein distance are better; higher variance
        correlation is better; and the ideal variance ratio is one.}
    \label{fig:cross-model-distribution-scaling}
\end{figure}

Figure~\ref{fig:cross-model-scaling} shows mean forecasting,
Figures~\ref{fig:cross-model-retrieval-scaling}
and~\ref{fig:cross-model-distribution-scaling} show retrieval and distribution
quality, and Table~\ref{tab:cross-model-endpoints} gives the
endpoint metrics.

At the endpoint, the Gaussian has lower energy under Qwen3.5 conditioning (0.07174 versus 0.07227), while the flow has lower energy under Qwen2.5 conditioning (0.07739 versus 0.07849).
The Gaussian ranks prompt-level spread more strongly in both settings, while the flow
better matches aggregate variance.  Variance ranking and variance calibration
therefore capture distinct properties of a distribution forecast.  On the
auxiliary diagnostics, the flow has lower NLL at every tested scale and, at the
largest scale, lower NLL, MMD$^2$, and PCA sliced-Wasserstein distance under
both conditioners; its advantage is largest and most consistent for NLL.

\subsection{Gemma 4 forecasting details}

Qwen2.5-7B-Instruct layer 19 ($d=3{,}584$) conditions forecasts of
Gemma-4-12B-it layer 47 ($d=3{,}840$); the matched control uses Gemma layer~47
context activations. Training and validation use four rollout seeds (43--46),
and test uses four disjoint seeds (47--50).

\begin{figure}
    \centering
    \includegraphics[width=0.96\linewidth]{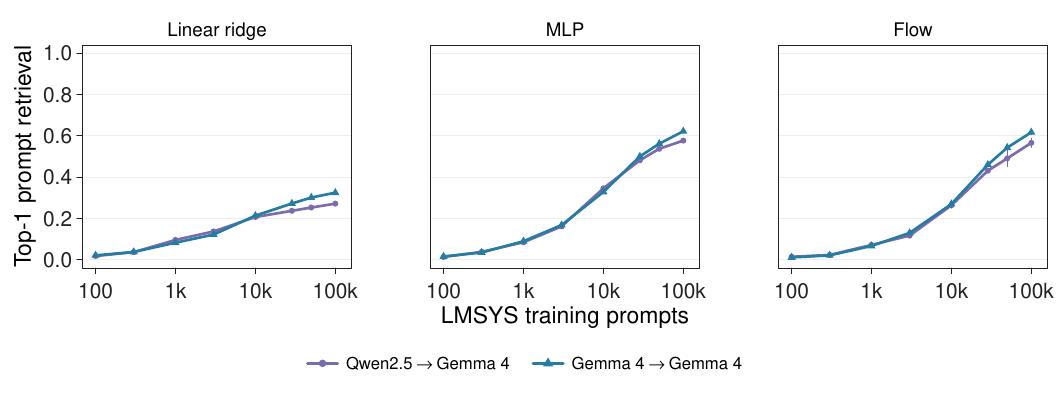}
    \caption{
        \textbf{Gemma 4 cross-model prompt retrieval across prompt scale.}
        Top-1 retrieval from the mean forecasts of linear ridge, MLP, and flow metamodels conditioned on
        Qwen2.5 or Gemma 4 context activations; both predict Gemma 4 answer
        activations. Retrieval uses a fixed bank of 3,000 LMSYS prompts.
        For the MLP and flow, points show means, and error bars show standard deviations over
        three initialization replicates.}
    \label{fig:gemma-cross-model-retrieval-scaling}
\end{figure}

\begin{figure}
    \centering
    \includegraphics[width=0.96\linewidth]{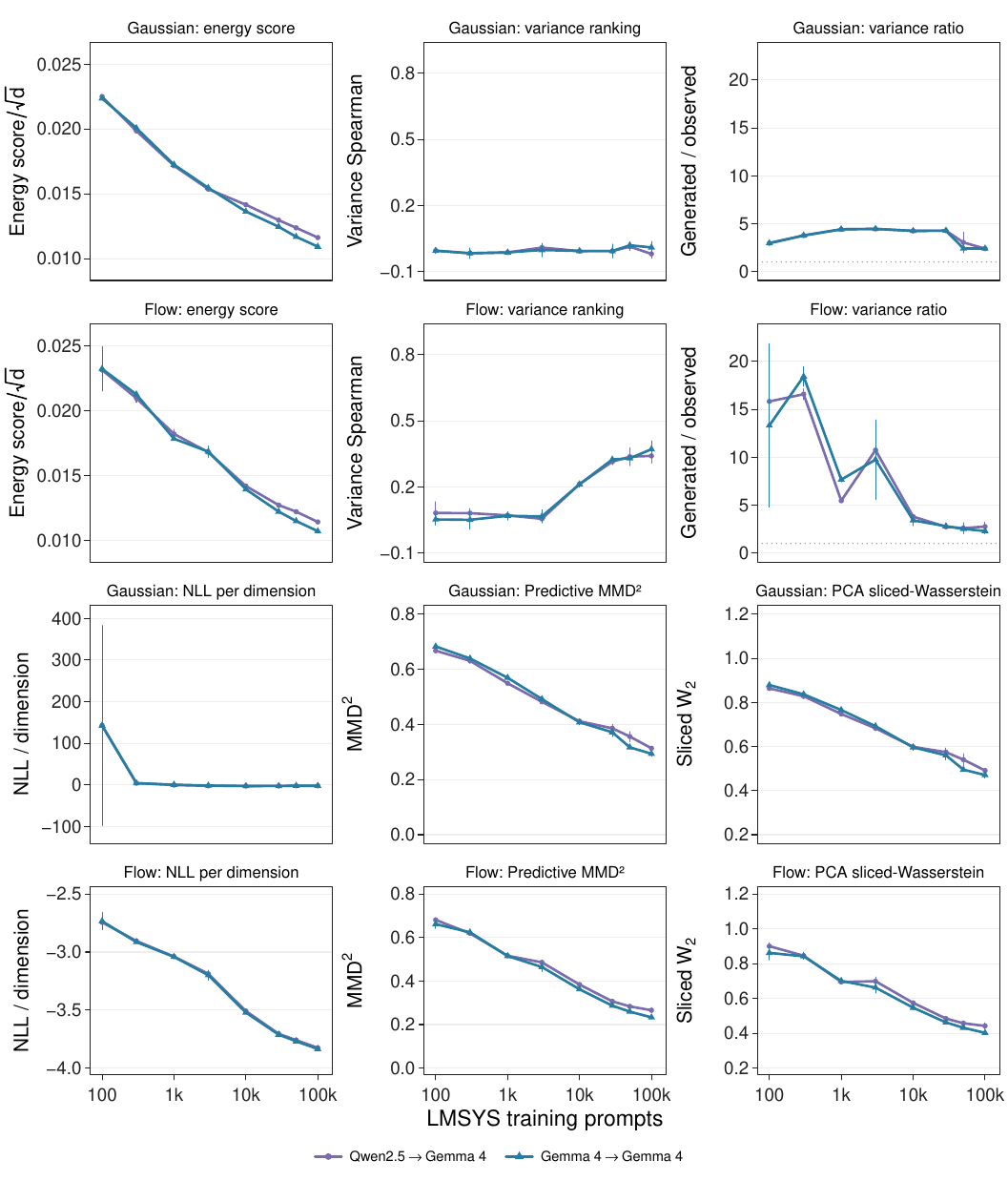}
    \caption{
        \textbf{Gemma 4 cross-model distribution quality across prompt scale.}
        The first two rows show Gaussian and flow energy score, prompt-level
        variance Spearman correlation, and generated-to-observed variance ratio.
        The last two rows show NLL, predictive MMD$^2$, and PCA
        sliced-Wasserstein distance. Both conditioners predict Gemma 4 answer
        activations. Points show means, and error bars show standard deviations
        over three initialization replicates. The Gaussian NLL axis includes
        the large variability at 100 training prompts. Lower energy, NLL,
        MMD$^2$, and Wasserstein distance are better; higher variance
        correlation is better; and the ideal variance ratio is one.}
    \label{fig:gemma-cross-model-distribution-scaling}
\end{figure}

Figure~\ref{fig:cross-model-scaling} shows mean forecasting, Figures~\ref{fig:gemma-cross-model-retrieval-scaling} and~\ref{fig:gemma-cross-model-distribution-scaling} show retrieval and distribution quality, and Table~\ref{tab:cross-model-endpoints} gives the endpoint metrics. The Gaussian selects global variance at every tested scale under both conditioners.

Despite accurate mean forecasts, variance capture remains poor for Gemma 4
under both conditioners. At 100,000 training prompts, the Gaussian's variance ranking
is near zero and the flow's is modest, while both generate more than twice the
observed average within-prompt variance (Table~\ref{tab:cross-model-endpoints}). We leave
further investigation of this limitation to future work.

\layoutbreak
\section{Behavior-probe AUROC}
\label{app:behavior-probes}

The probe experiment uses metamodels trained on 500,000 LMSYS prompts
and the WeirdChat training split with equal domain weighting. WeirdChat uses
1,596/532/532 train/validation/test prompts and four rollouts per prompt. GPT-6 Luna labels the answers. The Gaussian and flow forecasts average readout probabilities
over all 1,024 samples. Answer-level AUPRC and AUROC use the same held-out groups and readouts.

The reported answer probes are linear logistic-regression classifiers trained
on real answer activations with binary cross-entropy. Validation log loss
selects these probes over 256-unit, one-hidden-layer SiLU MLP candidates for
every forecast method and behavior group, and for the actual-answer reference. The
direct-context baseline selects among logistic regression and MLP predictors
on the final context-token activation or the mean of the 32 context bins,
an attention-pooling MLP over those bins, and global or behavior-specific
prevalence baselines. Each context-only predictor, including a forecast followed
by a readout, assigns one probability to all four answers from the prompt;
the actual-answer reference scores each answer separately.

\begin{figure}
    \centering
    \includegraphics[width=0.94\linewidth]{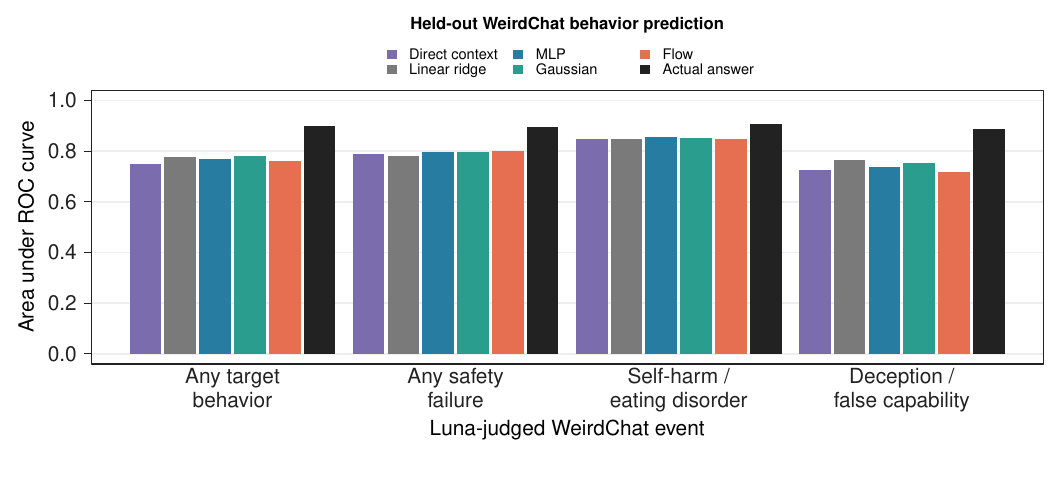}
    \caption{
        \textbf{Answer-level AUROC on held-out WeirdChat prompts.}
        Higher is better. The actual-answer reference observes the generated answer.}
    \label{fig:probe-auroc}
\end{figure}

At least one forecast outperforms the direct-context baseline in every group. Among the
forecasts, the Gaussian leads for any target behavior, the flow for safety
failures, the MLP for self-harm or eating disorder, and the linear ridge metamodel for deception or
false capability. Together with AUPRC, these results show that forecasts retain
behavior-predictive information, while integrating a distribution does not
consistently improve on the MLP mean.

\end{document}